\documentclass[conference]{IEEEtran}
\IEEEoverridecommandlockouts
\usepackage{cite}
\usepackage{amsmath,amssymb,amsfonts}
\usepackage{algorithmic}
\usepackage{graphicx}
\usepackage{textcomp}
\usepackage{xcolor}
\usepackage[utf8]{inputenc}
\usepackage{textgreek}
\usepackage{hyperref}
\usepackage{epstopdf}
\usepackage{subcaption} 
\usepackage{enumitem}
\usepackage{cuted}  

\graphicspath{{image/}}
\def\BibTeX{{\rm B\kern-.05em{\sc i\kern-.025em b}\kern-.08em
    T\kern-.1667em\lower.7ex\hbox{E}\kern-.125emX}}
\begin{document}

\title{Control Your Robot: A Unified System for Robot Control and Policy Deployment\\
}

\author{
\IEEEauthorblockN{Tian Nian\textsuperscript{1}*,
Weijie Ke\textsuperscript{2}*,
Shaolong Zhu\textsuperscript{3},
and Bingshan Hu\textsuperscript{2}\textsuperscript{†}}
\IEEEauthorblockA{\textsuperscript{1}ScaleLab, Shanghai Jiao Tong University  \textsuperscript{2}University of Shanghai for Science and Technology  \textsuperscript{3}Lumina Group }
\IEEEauthorblockA{*Equal contribution, †Corresponding author}
}

\maketitle

\begin{figure*}[!t]
  \centering
  \includegraphics[width=0.85\textwidth]{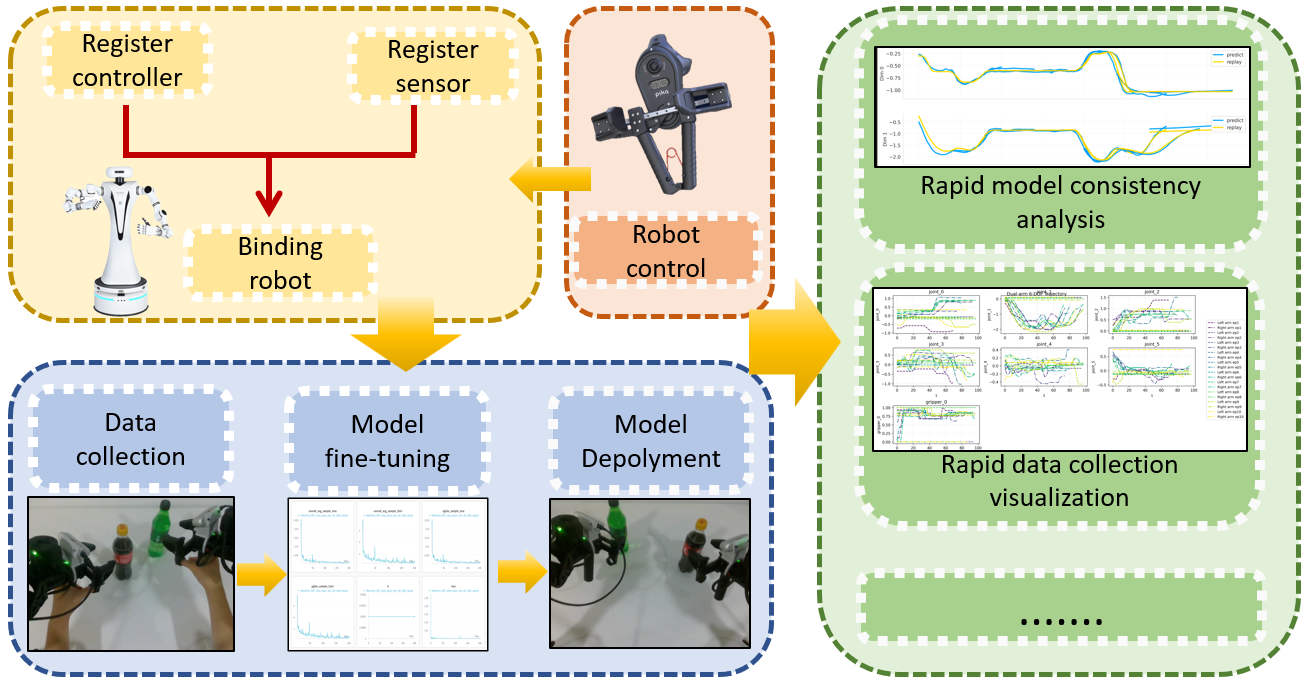}
  \caption{\textbf{System Design.} ``Control Your Robot'' provides a unified workflow that integrates robot control through controller and sensor registration. It enables seamless data collection, model training, and deployment on real hardware with diverse teleoperation devices. The system also includes data-analysis plugins (rapid visualization and efficient offline evaluation) for continuous assessment and improvement.}
  \label{fig:system}
\end{figure*}

\begin{abstract}
Cross-platform robot control remains difficult because hardware interfaces, data formats, and control paradigms vary widely, which fragments toolchains and slows deployment. To address this, we present \textbf{Control Your Robot}, a modular, general-purpose framework that unifies data collection and policy deployment across diverse platforms. The system reduces fragmentation through a standardized workflow with modular design, unified APIs, and a closed-loop architecture. It supports flexible robot registration, dual-mode control with teleoperation and trajectory playback, and seamless integration from multimodal data acquisition to inference. Experiments on single-arm and dual-arm systems show efficient, low-latency data collection and effective support for policy learning with imitation learning and vision-language-action models. Policies trained on data gathered by Control Your Robot match expert demonstrations closely, indicating that the framework enables scalable and reproducible robot learning across platforms.
\end{abstract}


\section{Introduction}

Recent progress in data-driven embodied-intelligence models has markedly advanced robotic control~\cite{kawaharazuka2025vision,long2025survey,mu2025robotwin,chen2025robotwin,chen2025benchmarking,liu2025avr,fu2025cordvip,motoda2025recipe}, establishing embodied intelligence as a central research paradigm. These gains are enabled by the availability of large-scale robot-manipulation datasets and the adoption of foundation models. Representative methods, including ACT~\cite{fu2024mobile}, G3Flow~\cite{chen2025g3flow}, Diffusion Policy~\cite{chi2023diffusion}, ManiCM~\cite{lu2024manicm}, OpenVLA~\cite{kim2024openvla}, RDT~\cite{liu2024rdt}, and $\pi_0$~\cite{black2024pi_0}, learn robust perceptual and control representations from human demonstrations and diverse corpora, and consequently achieve strong generalization across tasks, objects, and embodiments.

However, a significant deployment bottleneck persists for embodied intelligence models. Their adaptation to novel environments demands extensive, targeted data collection and fine-tuning, leading to substantial costs and system complexity. To overcome this limitation, we propose Control Your Robot, a unified and general-purpose framework designed to simplify the entire pipeline from cross-platform data collection to model fine-tuning and final deployment. Experimental results across various physical robot platforms demonstrate that our framework enables models to achieve highly consistent trajectory performance, confirming the effectiveness of the integrated workflow.
The main contributions of this work are summarized as follows:

\begin{enumerate}[leftmargin=1.5em,labelsep=0.5em]
    \item We introduce Control Your Robot, a modular, unified-interface, and closed-loop data collection–fine-tuning–deployment system designed to facilitate demonstration data collection across heterogeneous robotic platforms. 
    \item We experimentally demonstrate that Control Your Robot significantly improves the reusability of data collection systems.
\end{enumerate}
Control Your Robot is released as a fully open-source framework, providing APIs for data collection as well as training and deployment code. Control Your Robot available at \href{https://github.com/Tian-Nian/control_your_robot}{https://control-your-robot}.

\section{RELATED WORK}

\subsection{Robot Data Collection System}

Teleoperation remains a practical approach for collecting demonstrations, particularly in contact-rich tasks. \emph{SharedAssembly}~\cite{sharedassembly2025} enables efficient tele-assembly, while \emph{PATO}~\cite{pato2023} employs policy assistance to reduce operator effort and scale across robots. Platforms like \emph{RAPID Hand}~\cite{rapidhand2025} integrate visual, proprioceptive, and tactile sensing for fine-grained manipulation. Building on these principles, our framework provides a modular, low-latency pipeline for scalable demonstration capture and training.

\subsection{Robot Data Storage System}

Efficient storage is critical for reproducibility and scalability. \emph{LeRobot}~\cite{lerobot2024} offers a unified format with high-throughput access, while large-scale efforts like \emph{RT-X}~\cite{rtx2023} enable foundation-level learning over millions of trajectories. Such frameworks address indexing, compression, and retrieval. Our system complements these by providing a streamlined interface between collection and long-term management.

\subsection{General Embodied Model Deployment}

Deploying embodied models requires real-time perception–action loops and robust integration with dynamic environments. Existing frameworks typically separate inference, sensing, and control, often leveraging middleware like ROS~\cite{ros-hub}. Standardized interfaces and modular designs support policy transfer across robots and tasks~\cite{rai2025,ros-x-habitat2021}. Recent systems, such as RAI~\cite{rai2025} and ROS-LLM~\cite{ros-llm2024}, extend this paradigm by integrating large language models and structured reasoning for intuitive task programming. Similarly, ROS-X-Habitat~\cite{ros-x-habitat2021} bridges ROS and Habitat for seamless simulation–real transfer. Building on these ideas, our system offers a modular deployment interface aligned with data collection and storage, enabling reproducible real-world evaluation.

\section{System Architecture}

\subsection{Design Goals and Principles}
The proposed framework is guided by three design principles: \textit{modularity}, \textit{interface unification}, and \textit{process standardization}. 
These principles aim to address two persistent challenges in embodied intelligence research: fragmented data formats and heterogeneous hardware interfaces. 
By abstracting common functionalities and enforcing standardized workflows, the system lowers the barrier for deploying vision-language-action (VLA) models across diverse robotic platforms. 
In contrast to prior solutions that are often tailored to specific devices or data conventions, our design emphasizes generality, extensibility, and reproducibility.

\subsection{Overall Framework}
As illustrated in Fig.~\ref{fig:system}, the framework consists of three tightly integrated components:  
\begin{itemize}
    \item \textbf{Robot Registration}: A standardized interface layer for registering controllers and sensors, enabling seamless integration of new hardware modules.  
    \item \textbf{Robot Control}: A dual-mode control mechanism supporting both checkpoint-based trajectory execution and teleoperation-based interactive control, thereby ensuring precision while retaining operational flexibility.  
    \item \textbf{Data-to-Deployment Pipeline}: A closed-loop workflow that encompasses multi-modal data acquisition, unified data formatting, model fine-tuning, and inference deployment, bridging perception and action in a systematic manner.  
\end{itemize}

This architecture ensures that data generation, management, and model deployment form a coherent pipeline, facilitating efficient adaptation from controlled experimental setups to real-world robotic environments.
\subsection{System Details}

\subsubsection{Modularity}
The framework adopts a modular design that decomposes robotic systems into three fundamental units: \textit{controllers}, \textit{sensors}, and \textit{data processing modules}. 
Each unit exposes configurable and callable interfaces, allowing flexible assembly of complete systems (e.g., manipulator, gripper, and vision sensors) without modification to other components. 
This modularity guarantees extensibility and accelerates adaptation to new hardware environments.

\subsubsection{Interface Unification}
To accommodate diverse robotic platforms, the framework defines a minimal set of unified APIs. 
Developers implement these APIs once for each hardware platform, which then enables system-wide compatibility for both control and data acquisition. 
This abstraction eliminates the need to directly handle device-specific communication protocols, thereby significantly reducing cross-platform integration costs.

\subsubsection{Process Standardization}
The framework enforces a standardized pipeline that spans the entire lifecycle of embodied learning: \textit{data collection $\rightarrow$ preprocessing $\rightarrow$ model fine-tuning $\rightarrow$ inference deployment}. 
During data collection, multi-modal observations are synchronously captured and stored. 
Fine-tuning is facilitated through automated scripts and configuration templates for widely used VLA models. 
Deployment is achieved via a unified inference\\\\

\begin{strip}
\centering
\includegraphics[width=1.0\linewidth]{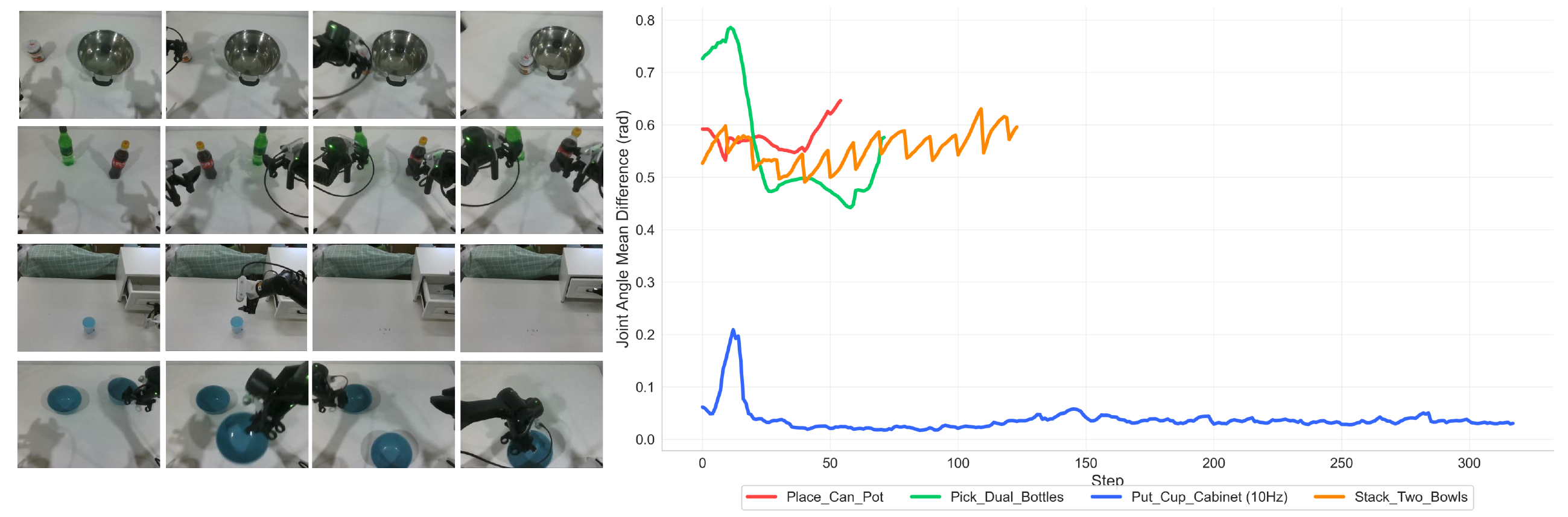}
\captionof{figure}{\textbf{Left:} We assess distinct manipulation tasks: \textbf{Place Can on Pot} (place a can on a pot edge); \textbf{Pick Dual Bottles} (simultaneously grasp two irregular bottles); \textbf{Put Cup in Cabinet} (handle recognition, drawer operation, placement, and closure); \textbf{Stack Two Bowls} (sequential pick-and-stack).\textbf{Right:}  The  section depicts the mean difference curves of action trajectories from the ACT model, obtained by replaying the trained model on the same tasks, aiming to evaluate its prediction accuracy.}
\label{fig:data}
\end{strip}

\noindent API, ensuring consistency between experimental evaluation and real-world execution.

This standardized process not only enhances system reusability but also improves reproducibility across experiments and platforms.

\section{experiment}


In this section, we evaluate the effectiveness of the proposed Control Your Robot framework. Specifically, we aim to address the following research questions:
\begin{itemize}
    \item How does "Control Your Robot" systematically contribute to policy robustness through enhanced data reliability?
    \item Can heterogeneous robotic systems based on Control Your Robot efficiently and accurately collect data?  
     
\end{itemize}

\subsection{Implementation Details and Task Description}
We conducted experiments on real-world robotic platforms, including both single-arm and dual-arm systems, across four representative manipulation tasks (see Fig.~\ref{fig:data} for real-world demonstrations).
 The single-arm robotic system consists of a slave DR-ALOHA arm and a master AGILEX-PIPER-ALOHA The slave arm is equipped with a wrist-mounted camera and a global camera, together forming a single-arm teleoperation system. The dual-arm robotic system consists of two IMETA-Y1-ALOHA arms configured in a master–slave integrated manner. Each arm is equipped with a wrist-mounted camera, and a shared global camera is positioned between the two arms, together forming a tabletop dual-arm teleoperation system.

All sensor data and robot state information were collected at 30 Hz. For policy learning, we employed ACT (a widely used imitation learning algorithm) and PI0 (an advanced vision-language-action model). For ACT training, we used the AdamW optimizer for 60k iterations. For PI fine-tuning, we applied a cosine learning rate scheduler for 30k iterations. Unless otherwise specified, all experiments were evaluated over 50 test trials for success rate.
Our method was assessed on a diverse set of tasks, each designed to test distinct robotic skill sets:
\begin{itemize}
    \item \textbf{Place Can Pot}: A spatial coordination task where the robot moves a sauce can from the table to a precise position next to the pot, requiring accurate placement and smooth motion control.
    \item \textbf{Pick Dual Bottles}: A dual-arm coordination task where each arm grasps a separate drink bottle simultaneously, emphasizing balance, synchronization, and grasp stability.
    \item \textbf{Put Cup Cabinet}: A hinge-based manipulation task in which the robot pulls open a sliding drawer, places a cup inside, and closes the drawer.
    \item \textbf{Stack Two Bowls}: A high-precision, long-horizon manipulation task that requires the model to accurately stack three bowls in the center of the workspace, following a strict order: right before left, and near before far.
\end{itemize}

Representative examples of each task are illustrated in the accompanying figures.
\subsection{Low-Latency Data Collection}
We additionally tested data acquisition with multiple sampling frequencies: the robotic arm was sampled at 600Hz, and images at 60Hz, using a dual-arm, three-view setup. The resulting effective sampling rates were 599.247Hz for the robotic arm and 59.997Hz for images, demonstrating that our system can reliably handle mixed-frequency data collection.

We also evaluated a multi-frequency control setup that combines high-frequency teleoperation with low-frequency data collection. Specifically, we used 300 Hz control for master–slave arm synchronization and 60 Hz for recording robot states and real-time images (the latter limited by the camera’s maximum frame rate). Over 10 continuous runs, the average real synchronization frequency stabilized at 282.35 Hz, and the data collection frequency at 59.99 Hz.

\subsection{Model Deployment and Data Replay}
During the deployment testing phase, 50 operational trajectories were executed and quantitatively compared with expert demonstration data. The results indicate a high degree of consistency between the generated trajectories and the expert data, with both variance and mean deviation maintained at low levels. In the data replay phase, both the ACT and PI models were trained using 20 sampled sequences (note: the PI model was excluded from graphical comparisons as it was trained on only a single task). Both models successfully replicated the expert operational strategies, as shown in (Fig.~\ref{fig:data}). With the exception of the \textbf{Put Cup Cabinet} task, the actual robot trajectories for the remaining tasks exhibited significant stochasticity, thereby imposing greater demands on the models' generalization capability. Owing to its limited parameter size, the ACT model demonstrated insufficient learning performance on real robot data, leading to notable performance deviations. Overall, the experimental findings confirm that the data collected by the "Control Your Robot" system effectively supports fine-grained policy learning.




\section{CONCLUSION}

In this paper, we presents Control Your Robot, a general-purpose robotic control system that offers a unified framework for data collection and model deployment. Comparative experiments demonstrate the efficacy of its data collection module and its versatility and scalability in both remote and local deployment scenarios. The system currently supports various mainstream robotic arms and ROS1/ROS2-based communication. Future work will focus on expanding hardware compatibility and integrating training/deployment functionalities for more control policies, aiming to develop this platform into a comprehensive tool for advancing robotic research and applications.


\bibliographystyle{IEEEtran}
\bibliography{references}

\end{document}